\documentclass[letter]{article}
\usepackage{proceed2e}
\usepackage{amssymb,amsfonts,amsbsy,amsmath,amsthm}
\usepackage{graphicx,subfigure}
\usepackage{enumitem}
\usepackage{multirow}
\usepackage{ifpdf}

\ifpdf
\usepackage[%
  pdftitle={Markov Chains on Orbits of Permutation Groups},%
  pdfauthor={Mathias Niepert},%
  pdfsubject={Conference on Uncertainty in Artificial Intelligence},%
  pdfkeywords={Markov chains, MCMC, lifted MCMC, lifted inference, symmetry, orbital Markov chains},%
  pdfstartview=FitH,%
  bookmarks=false,%
  bookmarksopen=false,%
  breaklinks=true,%
  colorlinks=true,%
  linkcolor=blue,anchorcolor=blue,%
  citecolor=blue,filecolor=blue,%
  menucolor=blue,pagecolor=blue,%
  urlcolor=blue]{hyperref}
\else
\usepackage[%
  breaklinks=true,%
  colorlinks=true,%
  linkcolor=blue,anchorcolor=blue,%
  citecolor=blue,filecolor=blue,%
  menucolor=blue,pagecolor=blue,%
  urlcolor=blue]{hyperref}
\fi

\theoremstyle{plain}
\newtheorem{thm}{Theorem}[section]
\newtheorem{theorem}[thm]{Theorem}

\newtheorem{corollary}[thm]{Corollary}

\theoremstyle{definition}

\newtheorem{example}[thm]{Example}

\title{Markov Chains on Orbits of Permutation Groups} 

\author{Mathias Niepert\\Universit\"{a}t Mannheim\\mniepert@gmail.com} 

\begin{document} 
 
\maketitle 
 
\begin{abstract} 
We present a novel approach to detecting and utilizing symmetries in probabilistic graphical models with two main contributions. First, we present a scalable approach to computing generating sets of permutation groups representing the symmetries of graphical models. Second, we introduce orbital Markov chains, a novel family of Markov chains leveraging model symmetries to reduce mixing times. We establish an insightful connection between model symmetries and rapid mixing of orbital Markov chains. Thus, we present the first lifted MCMC algorithm for probabilistic graphical models.  Both analytical and empirical results demonstrate the effectiveness and efficiency of the approach.
\end{abstract} 
 
\section{Introduction}

Numerous algorithms exploit model symmetries with the goal of reducing  the complexity of the computational problems at hand. Examples are procedures for detecting symmetries of first-order theories~\cite{crawford:1992} and propositional formulas~\cite{aloul:2003} in order to avoid the exhaustive exploration of a partially symmetric search space. More recently, symmetry detection approaches have been applied to answer set programming~\cite{drescher:2011} and (integer) linear programming~\cite{Margot:2003,Margot:2010,ostrowski:2011,mladenov12aistats}.
A considerable amount of attention to approaches utilizing model symmetries has been given by work on ``lifted probabilistic inference~\cite{poole:2003,braz:2005}." Lifted inference is mainly motivated by the large graphical models resulting from statistical relational formalism such as Markov logic networks~\cite{RD:2006}. The unifying theme of lifted probabilistic inference is that inference on the level of instantiated formulas is avoided and instead lifted to the first-order level. Notable approaches are lifted belief propagation~\cite{singla2008lifted,kersting2009counting}, bisimulation-based approximate inference algorithms~\cite{sen:2009}, first-order knowledge compilation techniques~\cite{broeck:2011,gogate2011probabilistic}, and lifted importance sampling approaches~\cite{gogate:2012}. With the exception of some  results for restricted model classes~\cite{singla2008lifted,broeck:2011,jaeger:2012}, there is a somewhat superficial understanding of the underlying principles of graphical model symmetries and the probabilistic inference algorithms utilizing such symmetries. Moreover, since most of the existing approaches are designed for relational models, the applicability to other types of probabilistic graphical models is limited.

The presented work contributes to a deeper understanding of the interaction between model symmetries and the complexity of inference by establishing a link between the degree of symmetry in graphical models and polynomial approximability. We describe the construction of colored graphs whose automorphism groups are equivalent to those of the graphical models under consideration.
We then introduce  the main contribution, \emph{orbital Markov chains}, the first general class of Markov chains for lifted inference. Orbital Markov chains combine the compact representation of symmetries with generating sets of permutation groups with highly efficient product replacement algorithms.
The link between model symmetries and polynomial mixing times of orbital Markov chains is established via a path coupling argument that is constructed so as to make the coupled chains coalesce whenever their respective states are located in the same equivalence class of the state space. The  coupling argument applied to orbital Markov chains opens up novel possibilities of analytically investigating classes of symmetries that lead to polynomial mixing times.

Complementing the analytical insights, we demonstrate empirically that orbital Markov chains converge faster to the true distribution than state of the art Markov chains on well-motivated and established sampling problems such as the problem of sampling independent sets from graphs. We also show that existing graph automorphism algorithms are applicable to compute symmetries of very large graphical models.

\section{Background and Related Work} 

We begin by recalling some basic concepts of group theory and finite Markov chains  both of which are crucial for understanding the presented work. In addition, we  give a brief overview of related work utilizing symmetries for the design of algorithms for logical and probabilistic inference.

\subsection{Group Theory}

A symmetry of a discrete object is a structure-preserving bijection on its components. For instance, a symmetry of a graph is a graph automorphism. Symmetries are often represented with permutation groups.  A group is an abstract algebraic structure ($\mathfrak{G}, \circ$), where $\mathfrak{G}$ is a set closed under a binary associative operation $\circ$ such that there is a identity element and every element has a unique inverse. Often, we refer to the group $\mathfrak{G}$ rather than to the structure ($\mathfrak{G}, \circ$). We denote the size of a group $\mathfrak{G}$ as $|\mathfrak{G}|$. A permutation group \emph{acting on} a finite set $\Omega$ is a finite set of bijections  $\mathfrak{g} : \Omega \rightarrow \Omega$ that form a group.

Let $\Omega$ be a finite set and let $\mathfrak{G}$ be a permutation group acting on $\Omega$. If $\alpha \in \Omega$ and $\mathfrak{g} \in \mathfrak{G}$ we write $\alpha^\mathfrak{g}$ to denote the image of $\alpha$ under $\mathfrak{g}$. 
A cycle $(\alpha_1\ \alpha_2\ ...\ \alpha_n)$
represents the permutation that maps $\alpha_1$ to $\alpha_2$, $\alpha_2$
to $\alpha_3$,..., and $\alpha_n$ to $\alpha_1$.
Every permutation can be written as a product of disjoint cycles where each element that does not occur in a cycle is understood as being mapped to itself. 
We define a relation $\sim$ on $\Omega$ with $\alpha \sim \beta$ if and only if there is a permutation $\mathfrak{g} \in \mathfrak{G}$ such that $\alpha^\mathfrak{g} = \beta$. The relation partitions $\Omega$ into equivalence classes which we call \emph{orbits}. We use the notation $\alpha^{\mathfrak{G}}$ to denote the orbit $\{\alpha^\mathfrak{g}\ |\ \mathfrak{g} \in \mathfrak{G}\}$ containing $\alpha$. 
Let $f: \Omega \rightarrow \mathbb{R}$ be a function from $\Omega$ into the real numbers and let $\mathfrak{G}$ be a permutation group acting on $\Omega$. 
We say that $\mathfrak{G}$ is an \emph{automorphism group for} $(\Omega, f)$ if and only if for all $\omega \in \Omega$ and all $\mathfrak{g} \in \mathfrak{G}$,  $f(\omega) = f(\omega^\mathfrak{g})$.

\subsection{Finite Markov chains}

Given a finite set $\Omega$ a \emph{finite Markov chain} defines a random walk $(X_0, X_1, ...)$ on elements of $\Omega$ with the property that the conditional distribution of $X_{n+1}$  given ($X_0, X_1, ..., X_n)$ depends only on $X_n$. For all $x, y \in \Omega$ $P(x, y)$ is the chain's probability to transition from $x$ to $y$, and $P^t(x, y)=P^{t}_{x}(y)$ the probability of being in state $y$ after $t$ steps if the chain starts at $x$. A Markov chain is \emph{irreducible} if for all $x, y \in \Omega$ there exists a $t$ such that $P^t(x, y) > 0$ and \emph{aperiodic} if for all $x \in \Omega$, $\mathsf{gcd}\{t \geq 1\ |\ P^t(x, x) > 0\} = 1$. A chain that is both irreducible and aperiodic converges to its unique stationary distribution. 

The total variation distance $d_{\mathsf{tv}}$ of the Markov chain from its stationary distribution $\pi$ at time $t$ with initial state $x$ is defined by 
$$\ d_{\mathsf{tv}}(P^{t}_{x},\pi) = \frac{1}{2} \sum_{y \in \Omega} |P^{t}(x, y) - \pi(y)|.$$ For $\varepsilon > 0$, let $\tau_x(\varepsilon)$ denote the least value $T$ such that $d_{\mathsf{tv}}(P^{t}_{x},\pi) \leq \varepsilon$ for all $t \geq T$. The  \emph{mixing time} $\tau(\varepsilon)$ is defined by $\tau(\varepsilon) = \max\{\tau_x(\varepsilon)\ |\ x \in \Omega\}$. We say that a Markov chain is \emph{rapidly mixing} if the mixing time is bounded by a polynomial in $n$ and $\log(\varepsilon^{-1})$, where $n$ is the size of each configuration in $\Omega$.

\subsection{Symmetries in Logic and Probability}

Algorithms that leverage model symmetries to solve computationally challenging problems more efficiently exist in several fields. Most of the work is related to the computation of symmetry breaking predicates to improve SAT solver performance~\cite{crawford:1992,aloul:2003}. The construction of our symmetry detection approach is largely derived from that of symmetry detection in propositional theories~\cite{crawford:1992,aloul:2003}. 
More recently, similar symmetry detection approaches have been put to work for answer set programming~\cite{drescher:2011} and integer linear programming~\cite{ostrowski:2011}.
Poole introduced the notion of lifted probabilistic inference as a variation of variable elimination taking advantage of the symmetries in graphical models resulting from probabilistic relational formalisms~\cite{poole:2003}. Following Poole's work, several algorithms for lifted probabilistic inference were developed such as lifted and counting belief propagation~\cite{singla2008lifted,kersting2009counting}, bi-simulation-based approximate inference~\cite{sen:2009}, general purpose MCMC algorithm for relational models~\cite{milch:2006} and, more recently, first-order knowledge compilation techniques~\cite{broeck:2011,gogate2011probabilistic}.  In contrast to existing methods, we present an approach that is applicable to a much larger class of graphical models.

\section{Symmetries in Graphical Models}

Similar to the method of symmetry detection in propositional  formulas~\cite{crawford:1992,aloul:2003,darga:2008} we can, for a large class of probabilistic graphical models, construct a colored undirected graph whose automorphism group is equivalent to the permutation group representing the model's symmetries.  We describe the approach for sets of partially weighted propositional formulas since  Markov logic networks, factor graphs, and the weighted model counting framework can be represented using sets of (partially) weighted formulas~\cite{RD:2006,broeck:2011,gogate2011probabilistic}. 
\begin{figure}[t!]
\begin{center}
\includegraphics[width=0.3\textwidth]{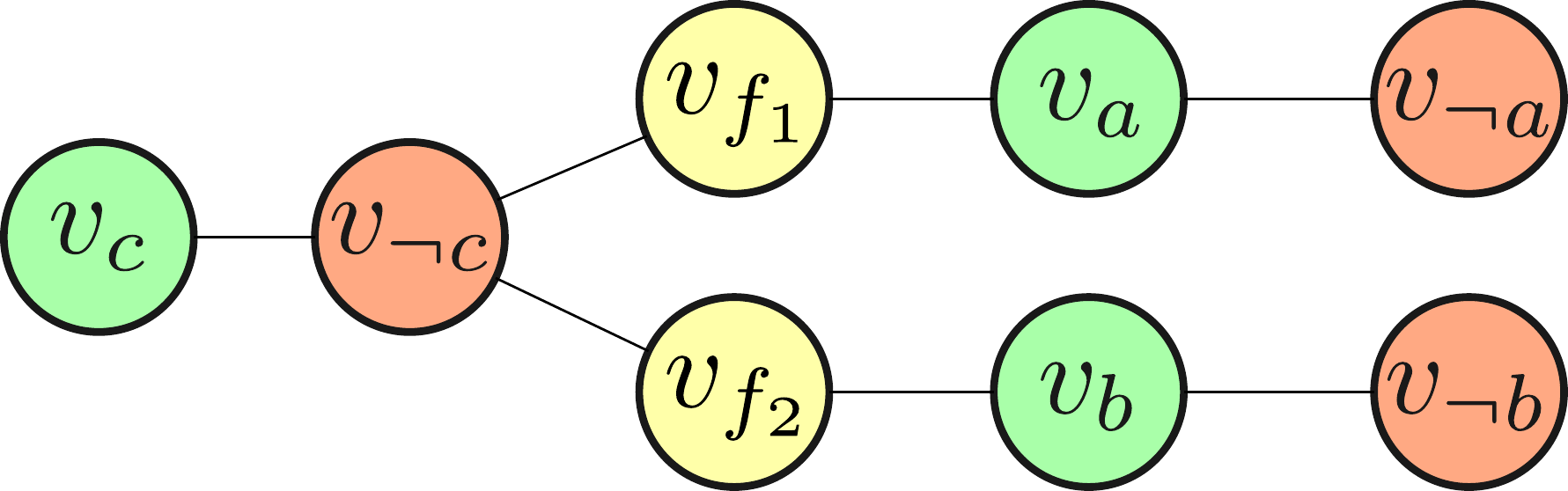}
\caption{\label{fig:cg}The colored graph resulting from the set of weighted clauses of Example~\ref{example-wcnf}.}
\end{center}
\end{figure}
For the sake of readability, we describe the colored graph construction for partially weighted \emph{clauses}. Using a more involved transformation, however, we can extend it to sets of partially weighted formulas. Let $\mathcal{S} = \{(f_i, w_i)\}, 1 \leq i \leq n,$ be a set of partially weighted clauses with $w_i \in \mathbb{R}$ if $f_i$ is weighted and $w_i=\infty$ otherwise. We define an \emph{automorphism of} $\mathcal{S}$ as a permutation mapping (a) unnegated variables to unnegated variables, (b) negated variables to negated variables, and (c) clauses to clauses, respectively, such that this permutation maps $\mathcal{S}$ to an identical set of partially weighted clauses. The set of these permutations forms the automorphism group of $\mathcal{S}$.

The construction of the colored undirected graph $G(\mathcal{S}) = (V,E)$ proceeds as follows. For each variable $a$ occurring in $\mathcal{S}$ we add two nodes $v_{a}$ and $v_{\neg a}$ modeling the unnegated and negated variable, respectively, to $V$ and the edge $\{v_{a}, v_{\neg a}\}$ to $E$. We assign color $0$ ($1$) to nodes corresponding to negated (unnegated) variables. This coloring precludes permutations that map a negated variable to an unnegated one or vice versa. We introduce a distinct color $c_{\infty}$ for unweighted clauses and a color $c_w$ for each distinct weight $w$ occurring in $\mathcal{S}$. For each clause $f_i$ with weight $w_i=w$ we add a node $v_{f_i}$ with color $c_w$ to $V$. For each unweighted clause $f_i$ we add a node $v_{f_i}$ with color $c_{\infty}$ to $V$. Finally, we add edges between each clause node $v_{f_i}$ and the nodes of the negated and unnegated variables occurring in $f_i$. Please note that we can incorporate evidence by introducing two novel and distinct colors representing \emph{true} and \emph{false} variable nodes.

\begin{example}
\label{example-wcnf}
Let $\{f_1:=(a \vee \neg c, 0.5), f_2:=(b \vee \neg c, 0.5)\}$ be a set of weighted clauses. We introduce $6$ variable nodes $v_{a}, v_{b}, v_{c}, v_{\neg a}, v_{\neg b}, v_{\neg c}$ where the former three have color $1$ (green) and the latter three color $0$ (red). We connect the nodes $v_{a}$ and $v_{\neg a}$; $v_{b}$ and $v_{\neg b}$; and $v_{c}$ and $v_{\neg c}$. We then introduce two new clause nodes $v_{f_1}, v_{f_2}$ both with color $2$ (yellow) since they have the same weight. We finally connect the variable nodes with the clause nodes they occur in.  Figure~\ref{fig:cg} depicts the resulting colored graph. A generating set of $\mathsf{Aut}(G(\mathcal{S}))$, the automorphism group of this particular colored graph, is $ \{(v_a\ v_b)(v_{\neg a}\ v_{\neg b})(v_{f_1}\ v_{f_2})\}$.
\end{example}

The following theorem states the relationship between the automorphisms of $\mathcal{S}$ and the colored graph $G(\mathcal{S})$.

\begin{theorem}
\label{theorem-automorphism}
 Let $\mathcal{S} = \{(f_i, w_i)\}, 1 \leq i \leq n,$ be a set of partially weighted clauses and let $\mathsf{Aut}(G(\mathcal{S}))$ be the automorphism group of the colored graph constructed for $\mathcal{S}$. There is a one-to-one correspondence between $\mathsf{Aut}(G(\mathcal{S}))$ and the automorphism group of $\mathcal{S}$. 
\end{theorem} 

Given a set of partially weighted clauses $\mathcal{S}$ with variables $\mathbf{X}$ we have, by Theorem~\ref{theorem-automorphism}, that if we define a distribution $\Pr$ over random variables $\mathbf{X}$  with features $f_i$ and  weights $w_i$, $1\leq i \leq n$, then  $\mathsf{Aut}(G(\mathcal{S}))$ is an automorphism group for $(\{0,1\}^{\mathbf{X}}, \Pr)$. Hence, we can use the method to find symmetries in a large class of graphical models.
The complexity of computing generating sets of $\mathsf{Aut}(G(\mathcal{S}))$ is in NP and not known to be in P or NP-complete. For graphs with bounded degree  the problem is in P~\cite{luks:1982}. There are specialized algorithms for finding generating sets of automorphism groups of colored graphs such as \textsc{Saucy}\cite{darga:2008} and \textsc{Nauty}\cite{mckay:1981} with remarkable performance. We will show that \textsc{Saucy} computes \emph{irredundant} sets of generators of automorphism groups for graphical models with millions of variables. The size of these generating sets is bounded by the number of graph vertices.

We briefly position the symmetry detection approach in the context of existing algorithms and concepts.

\subsection{Lifted Message Passing}

There are two different lifted message passing algorithms. Lifted First-Order Belief Propagation~\cite{singla2008lifted} operates on Markov logic networks whereas Counting Belief Propagation~\cite{kersting2009counting} operates on factor graphs. Both approaches leverage symmetries in the model to partition variables and features into equivalence classes. Each variable class (supernode/clusternode) contains those variable nodes that would send and receive the same messages were (loopy) belief propagation (BP) run on the original model. Each feature class (superfeature/clusterfactor) contains factor nodes that would send and receive the same BP messages.

The colored graph construction provides an alternative approach to partitioning the variables and features of a graphical model.  We simply compute the \emph{orbit partition} induced by the permutation group $\mathsf{Aut}(G(\mathcal{S}))$ acting on the set of variables and features. For instance, the orbit partition of Example~\ref{example-wcnf} is $\{\{a, b\}, \{c\}, \{f_1, f_2\}\}.$ In general, orbit partitions have the following properties: For two variables $v_1,v_2$ in the same orbit we have that (a) $v_1$ and $v_2$ have identical marginal probabilities and (b) the variable nodes corresponding to $v_1$ and $v_2$ would send and receive the same messages were BP run on the original model; and for two features $f_1$ and $f_2$ in the same orbit we have that the factor nodes corresponding to $f_1$ and $f_2$ would send and receive the same BP messages.

\subsection{Finite Partial Exchangeability} 
 
The notion of exchangeability was introduced by de Finetti~\cite{finetti:1972}. Several theorems concerning finite (partial) exchangeability have been stated~\cite{diaconis:1980,finetti:1972}. 
Given a finite sequence of $n$ binary random variables $\mathbf{X}$, we say that $\mathbf{X}$ is \emph{exchangeable} with respect to the distribution $\Pr$ if, for every $\mathbf{x} \in \{0,1\}^n$ and \emph{every} permutation $\mathfrak{g}$ acting on $\{0,1\}^n$, we have that $\Pr(\mathbf{X}=\mathbf{x}) = \Pr(\mathbf{X}=\mathbf{x}^\mathfrak{g}).$ This is equivalent to saying that the symmetric group $\mathsf{Sym}(n)$ is an automorphism group for $(\{0,1\}^n,\Pr)$. Whenever we have finite exchangeability, there are $n+1$ orbits each containing the variable assignments with Hamming weight $i, 0 \leq i \leq n$.   Hence, every exchangeable probability distribution over $n$ binary random variables is a unique mixture of draws from the $n+1$ orbits. In some cases of partial exchangeability, namely when the orbits can be specified using a \emph{statistic}, one can use this for a more compact representation of the distribution as a product of mixtures~\cite{diaconis:1980}. The symmetries that have to be present for such a re-parameterization to be feasible, however, are rare and constitute one end of the symmetry spectrum. 

Therefore, a central question is how \emph{arbitrary symmetries}, compactly represented with irredundant generators of permutation groups, can be utilized for  efficient probabilistic inference algorithms that go beyond (a) single variable marginal inference via lifted message passing and (b) the limited applicability of finite partial exchangeability. In order to answer this question, we turn to the major contribution of the present work.

\section{Orbital Markov Chains}

Inspired by the previous observations, we introduce \emph{orbital Markov chains}, a novel family of Markov chains. An orbital Markov chain is always derived from an existing Markov chain so as to leverage the symmetries in the underlying model. In the presence of symmetries orbital Markov chains are able to perform wide-ranging transitions reducing the time until convergence. In the absence of symmetries they are equivalent to the original Markov chains. Orbital Markov chains only require a generating set of a permutation group $\mathfrak{G}$ acting on the chain's state space as additional input. As we have seen, these sets of generators are computable with graph automorphism algorithms.

Let $\Omega$ be a finite set, let $\mathcal{M}' = (X'_0, X'_1, ...)$ be a Markov chain with state space $\Omega$, let $\pi$ be a stationary distribution of $\mathcal{M}'$, and let $\mathfrak{G}$ be an automorphism group for $(\Omega, \pi)$. The \emph{orbital Markov chain}  $\mathcal{M}=(X_0, X_1, ...)$ \emph{for} $\mathcal{M'}$ is a Markov chain which at each integer time $t+1$ performs the following steps:

\begin{enumerate}
\item Let $X'_{t+1}$ be the state of the original Markov chain $\mathcal{M}'$ at time $t+1$;
\item Sample $X_{t+1}$, the state of the orbital Markov chain $\mathcal{M}$ at time  $t+1$, uniformly at random from ${X'}_{t+1}^{\mathfrak{G}}$, the orbit of $X'_{t+1}$.
\end{enumerate}

The orbital Markov chain $\mathcal{M}$, therefore, runs at every time step $t\geq 1$ the original chain $\mathcal{M}'$ first and samples the state of $\mathcal{M}$ at time $t$  uniformly at random from the orbit of the state of the original chain $\mathcal{M'}$ at time $t$. 

First, let us analyze the complexity of the second step which differs from the original Markov chain. Given a state $X_{t}$ and a permutation group $\mathfrak{G}$ we need to sample an element from ${X_{t}}^{\mathfrak{G}}$, the orbit of $X_{t}$,  uniformly at random. By the orbit-stabilizer theorem this is equivalent to sampling an element $\mathfrak{g} \in \mathfrak{G}$ uniformly at random and computing ${X_{t}}^{\mathfrak{g}}$. Sampling group elements nearly uniform at random is a well-researched  problem~\cite{celler:1995} and computable in polynomial time in the size of the generating sets with product replacement algorithms~\cite{pak:2000}. These algorithms are implemented in several group algebra systems such as \textsc{Gap}\cite{GAP4} and exhibit remarkable performance.  Once initialized, product replacement algorithms can generate pseudo-random elements by performing  a small number of group multiplications. We could  verify that the overhead of step $2$  during the sampling process is indeed negligible.

Before we analyze the conditions under which orbital Markov chains are aperiodic, irreducible, and have the same stationary distribution as the original chain, we provide an example of an orbital Markov chain that is based on the standard Gibbs sampler which is commonly used to perform probabilistic inference.

\begin{example}
Let $\mathbf{V}$ be a finite set of random variables with  probability distribution $\pi$, and let $\mathfrak{G}$ be an automorphism group for $(\times_{V \in \mathbf{V}} V, \pi)$.
The orbital Markov chain for the \emph{Gibbs sampler} is a Markov chain $\mathcal{M}=(X_0, X_1, ...)$ which, being in state $X_t$, performs the following steps at time $t+1$:
\begin{enumerate}
\item Select a variable $V \in \mathbf{V}$ uniformly at random;
\item Sample $X'_{t+1}(V)$, the value of $V$ in the configuration $X'_{t+1}$, according to the conditional $\pi$-distribution of $V$ given that all other variables take their values according to ${X}_{t}$;
\item Let $X'_{t+1}(W)=X_{t}(W)$ for all variables $W \in \mathbf{V} \setminus \{V\}$; and
\item Sample $X_{t+1}$ from ${X'}^{\mathfrak{G}}_{t+1}$, the orbit of $X'_{t+1}$, uniformly at random.
\end{enumerate}
We call this Markov chain the \emph{orbital Gibbs sampler}. 
In the absence of  symmetries, that is, if $\mathfrak{G}$'s only element is the  identity permutation, the orbital Gibbs sampler is equivalent to the standard Gibbs sampler.
\end{example}

Let us now state a major result of this paper. It relates properties of the orbital Markov chain to those of the Markov chain it is derived from. A detailed proof can be found in the appendix.

\begin{theorem}
\label{theorem-orbital}
Let $\Omega$ be a finite set and let $\mathcal{M}'$ be a Markov chain with state space $\Omega$ and transition matrix $P'$. Moreover, let $\pi$ be a probability distribution on $\Omega$, let $\mathfrak{G}$ be an automorphism group for $(\Omega, \pi)$, and let $\mathcal{M}$ be the orbital Markov chain for $\mathcal{M'}$. Then,
\begin{itemize}
\item[(a)] if $\mathcal{M'}$ is aperiodic then $\mathcal{M}$ is also aperiodic;
\item[(b)] if $\mathcal{M'}$ is irreducible then $\mathcal{M}$ is also irreducible; 
\item[(c)] if $\pi$ is a reversible distribution for $\mathcal{M'}$ and, for all $\mathfrak{g} \in \mathfrak{G}$ and all $x, y \in \Omega$ we have that $P'(x, y)=P'(x^{\mathfrak{g}}, y^{\mathfrak{g}})$, then $\pi$ is also a reversible and, hence, a stationary distribution for $\mathcal{M}$.
\end{itemize}
\end{theorem}
The condition in statement (c) requiring for all $\mathfrak{g} \in \mathfrak{G}$ and all $x, y \in \Omega$ that $P'(x, y)=P'(x^{\mathfrak{g}}, y^{\mathfrak{g}})$ conveys that the original Markov chain is compatible with the symmetries captured by the permutation group $\mathfrak{G}$. This rather weak assumption is met by all of the practical Markov chains we are aware of and, in particular, Metropolis chains and the standard Gibbs sampler. 

\begin{corollary}
\label{corollary-orbital-gibbs}
Let $\mathcal{M'}$ be the Markov chain of the Gibbs sampler with reversible distribution $\pi$. The orbital Gibbs sampler for $\mathcal{M'}$ is aperiodic and has $\pi$ as a reversible and, hence, a stationary distribution. Moreover, if $\mathcal{M'}$ is irreducible then the orbital Gibbs sampler is also irreducible and it has $\pi$ as its unique stationary distribution.
\end{corollary}

\begin{figure}[t!]
\begin{center}
\includegraphics[width=0.35\textwidth]{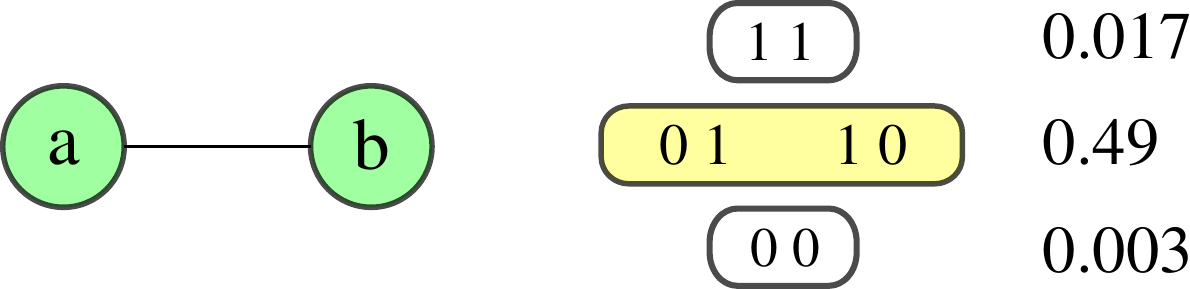}
\caption{\label{fig:example} An undirected graphical model over two binary variables and one symmetric potential function with corresponding distribution shown on the right. There are three orbits each indicated by one of the rounded rectangles.  }
\end{center}
\end{figure}

We will show both analytically and empirically that, in the presence of symmetries, the orbital Gibbs sampler converges at least as fast or faster to the true distribution than state of the art sampling algorithms. First, however, we want to take a look at an example that illustrates the advantages of the orbital Gibbs sampler. 

\begin{example}
Consider the undirected graphical model in Figure~\ref{fig:example} with two binary random variables and a symmetric potential function. The probabilities of the states $01$ and $10$ are both $0.49$. Due to the symmetry in the model, the states $10$ and $01$ are part of the same orbit. Now, let us assume a standard Gibbs sampler is in state $10$. The probability for it to transition to one of the states $11$ and $00$ is only $0.02$ and,  by definition of the standard Gibbs sampler, it cannot transition directly to the state $01$. The chain is ``stuck" in the state $10$ until it is able to move to $11$ or $00$. Now, consider the orbital Gibbs sampler. Intuitively, while it is ``waiting" to move to one of the low probability states, it samples the two high probability states horizontally  uniformly at random from the orbit $\{01, 10\}$. In this particular case the  orbital Gibbs sampler converges faster than the standard Gibbs sampler, a fact that we will also show analytically.
\end{example}

\subsection{Mixing Time of Orbital Markov Chains}

We will make our intuition about the faster convergence of the orbital Gibbs sampler  more concrete. We accomplish this by showing that the more symmetry there is in the model the faster a coupling of the orbital Markov chain will coalesce and, therefore, the faster the chain will converge to its stationary distribution.

There are several methods available to prove rapid mixing of a finite Markov chain. The method we will use here is that of a \emph{coupling}. A coupling for a Markov chain $\mathcal{M}$ is a stochastic process $(X_t, Y_t)$ on $\Omega \times \Omega$ such that $(X_t)$ and $(Y_t)$ considered marginally are \emph{faithful copies} of $\mathcal{M}$. The \emph{coupling lemma} expresses that the total variation distance of $\mathcal{M}$ at time $t$ is limited from above by the probability that the two chains have not \emph{coalesced}, that is, have not met at time $t$ (see for instance Aldous~\cite{aldous:1983}). Coupling proofs on the joint space $\Omega \times \Omega$ are often rather involved and require complex combinatorial arguments. A possible simplification is provided by the \emph{path coupling} method where a coupling is only required to hold on a subset of $\Omega \times \Omega$~(Bubley and Dyer~\cite{bubley:1997}). The following theorem formalizes this idea.

\begin{theorem}[Dyer and Greenhill~\cite{dyer:1998}]
Let $\delta$ be an integer valued metric defined on $\Omega \times \Omega$ taking values in $\{0, ..., D\}$. Let $S \subseteq \Omega \times \Omega$ such that for all ($X_t, Y_t) \in \Omega \times \Omega$ there exists a path $X_t = Z_0, ..., Z_r = Y_t$ between $X_t$ and $Y_t$ with $(Z_l, Z_{l+1}) \in S$ for $0 \leq l \leq r$ and $$\sum_{l=0}^{r - 1} \delta(Z_l, Z_{l+1}) = \delta(X_t, Y_t).$$ Define a coupling $(X, Y) \rightarrow (X', Y')$ of the Markov chain $\mathcal{M}$ on all pairs $(X,Y) \in S$. Suppose there exists $\beta \leq 1$ with $\mathbf{E}[\delta(X', Y')] \leq \beta \delta(X,Y)$ for all $(X, Y) \in S$. If $\beta < 1$ then the mixing time $\tau(\varepsilon)$ of $\mathcal{M}$ satisfies $$\tau(\varepsilon) \leq \frac{\ln(D \varepsilon^{-1})}{1-\beta}.$$ If $\beta=1$ and there exists an $\alpha > 0$ such that $\Pr[\delta(X_{t+1},Y_{t+1}) \neq \delta(X_t, Y_t)] \geq \alpha$ for all $t$, then $$\tau(\varepsilon) \leq \left\lceil \frac{e D^2}{\alpha}\right\rceil \left\lceil \ln(\varepsilon^{-1})\right\rceil.$$
\end{theorem} 

We selected the \emph{insert/delete} Markov chain for independent sets of graphs for our analysis. Sampling independent sets is a classical problem motivated by numerous applications and with a considerable amount of recent research devoted to it~\cite{luby:1999,dyer2000markov,weitz:2006,sly:2010,restrepo:2011}. The coupling proof for the orbital version of this Markov chain provides interesting insights into the construction of such a coupling and  the influence of the graph symmetries on the mixing time. The proof strategy is in essence applicable to other sampling algorithms.

Let $G = (V,E)$ be a graph. A subset $X$ of $V$ is an \emph{independent set} if $\{v, w\} \notin E$ for all $v, w \in X$. Let $\mathcal{I}(G)$ be the set of all independent sets in a given graph $G$ and let $\lambda$ be a positive real number. The partition function $Z = Z(\lambda)$ and the corresponding probability measure $\pi_{\lambda}$ on $\mathcal{I}(G)$ are defined by $$Z=Z(\lambda) = \sum_{X \in \mathcal{I}(G)} \lambda^{|X|} \mbox{ \ \ and \ \ } \pi_{\lambda}(X) = \frac{\lambda^{|X|}}{Z}.$$ Approximating the partition function and  sampling from $\mathcal{I}(G)$ can be accomplished using a rapidly mixing Markov chain with state space $\mathcal{I}(G)$ and stationary distribution $\pi_{\lambda}$. The simplest Markov chain for independent sets is the so-called \emph{insert/delete} chain~\cite{dyer2000markov}. If $X_t$ is the state at time $t$ then the state at time $t+1$ is determined by the following procedure:
\begin{enumerate}
\item Select a vertex $v \in V$ uniformly at random;
\item If $v \in X_t$ then let $X_{t+1} = X_t \setminus \{v\}$ with probability $1/(1 + \lambda)$;
\item If $v\notin X_t$ and $v$ has no neighbors in $X_t$ then let $X_{t+1}=X_{t} \cup \{v\}$ with probability $\lambda/(1+\lambda)$;
\item Otherwise let $X_{t+1} = X_{t}$.
\end{enumerate}

Using a path coupling argument one can show that the \emph{insert/delete} chain is rapidly mixing for $\lambda \leq 1/(\Delta - 1)$ where $\Delta$ is the maximum degree of the graph~\cite{dyer2000markov}. We can  turn the \emph{insert/delete} Markov chain into the orbital \emph{insert/delete} Markov chain $\mathcal{M}(\mathcal{I}(G))$ simply by adding the following fifth step:
\begin{enumerate}[resume]
\item Sample $X_{t+1}$ uniformly at random from its orbit.
\end{enumerate}

By Corollary~\ref{corollary-orbital-gibbs} the orbital \emph{insert/delete}  chain for independent sets is aperiodic, irreducible, and has $\pi_{\lambda}$ as its unique stationary distribution. We can now state the following theorem concerning the mixing time of this Markov chain. It relates the graph symmetries to the mixing time of the chain. The proof of the theorem is based on a path coupling that is constructed so as to make the two chains coalesce whenever their respective states are located in the same orbit. A detailed and instructive proof can be found in the appendix.

\begin{theorem}
\label{theorem-rho}
Let $G = (V,E)$ be a graph with maximum degree $\Delta$, let $\lambda$ be a positive real number, and let $\mathfrak{G}$ be an automorphism group for $(\{0,1\}^{V}, \pi_{\lambda})$. Moreover, let $(X\cup\{v\}) \in \mathcal{I}(G)$, let $(X\cup\{w\}) \in \mathcal{I}(G)$, let $\{v, w\} \in E$, and let $\rho = \Pr[(X\cup\{v\}) \notin (X\cup\{w\})^\mathfrak{G}].$ The orbital \emph{insert/delete} chain $\mathcal{M}(\mathcal{I}(G))$ is rapidly mixing if either $\rho \leq 0.5$ or $\lambda \leq 1/((2\rho - 1)\Delta - 1)$.
\end{theorem}

\begin{figure}[t!]
\begin{center}
\includegraphics[width=0.48\textwidth]{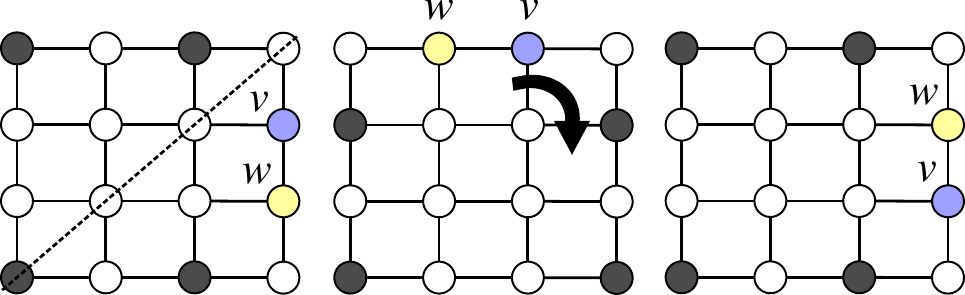}
\caption{\label{fig:hmodel_4x4}Two independent sets $X \cup\{v\}$ and $X \cup \{w\}$ of the $4\mathsf{x}4$ grid located in the same orbit. The first permutation is the reflection on the sketched diagonal the second a clockwise $90^{\circ}$ rotation.}
\end{center}
\end{figure}

The theorem establishes the important link between the graph automorphisms and the mixing time of the orbital \emph{insert/delete} chain. The more symmetries the graph exhibits the larger the orbits and the sooner the chains coalesce. Figure~\ref{fig:hmodel_4x4} depicts the $4\mathsf{x}4$ grid with two independent sets $X \cup\{v\}$ and $X \cup \{w\}$ with $\{v, w\} \in E$ and $(X \cup\{v\}) \in (X \cup \{w\})^\mathfrak{G}$. Since $\rho < 1$ for $n\mathsf{x}n$ grids, $n\geq 4$, we can prove (a) rapid mixing of the orbital \emph{insert/delete} chain for larger $\lambda$ values and (b) more rapid mixing for identical $\lambda$ values. 

The next corollary follows from Theorem~\ref{theorem-rho} and the simple fact that $X' \in X^\mathfrak{G}$ for all $X\subseteq V, X' \subseteq V$ with $|X|=|X'|$ whenever $\mathfrak{G}$ is the symmetric group on $V$.

\begin{corollary}
Let $G = (V, E)$ be a graph, let $\lambda$ be a positive real number, and let $\mathfrak{G}$ be an automorphism group for $(\{0,1\}^{V}, \pi_{\lambda})$. If $\mathfrak{G}$ is the symmetric group $\mathsf{Sym}(V)$ then $\mathcal{M}(\mathcal{I}(G))$ is rapidly mixing with $\tau(\varepsilon) \leq |V|\ln(|V| \varepsilon^{-1})$.
\end{corollary}

By analyzing the coupling proof of Theorem~\ref{theorem-rho} and, in particular, the moves leading to states $(X',Y')$ with $|X'|=|Y'|$  and the probability that $X'$ and $Y'$ are located in the same orbit in these cases, it is possible to provide more refined bounds.  Moreover, to capture the full power of orbital Markov chains, a coupling argument should not merely consider pairs of states with Hamming  distance $1$. Indeed, the strength of the orbital chains is that, in the presence of symmetries in the graph topology, there is a non-zero probability that states with large Hamming distance (up to $|V|$) are located in the same orbit. The method presented here is also applicable to Markov chains known to mix rapidly for larger $\lambda$ values than the \emph{insert/delete} chain such as the \emph{insert/delete/drag} chain~\cite{dyer2000markov}.

\begin{table}[t!]
\small
\begin{tabular}{|c|c|c|c|c|c|}
\hline
\multicolumn{6}{|c|}{social network model \cite{singla2008lifted}} \\ 
\hline 
people  & 20 & 50 & 100 & 250 & 500  \\ 
\hline
vertices & 1740 & 10200 & 40700 & 251750 & 1003500 \\
\hline
edges & 2120 & 10350 & 50600 & 314000 & 1253000 \\
\hline
time [s] & 0.04 & 0.15 & 0.81 & 22.5 & 261.3  \\ 
\hline
features & 860 & 5150 & 20300 & 125750 & 501500 \\
\hline
orbs w/o & 7 & 7 & 7 & 7 & 7 \\
\hline
orbs w/ & 238 & 1244 & 6237 & 30192 & 78303 \\
\hline 
\hline
\multicolumn{6}{|c|}{$k\mathsf{x}k$ grid model} \\ 
\hline
$k$ & 20 & 50 & 100 & 250 & 500 \\
\hline
vertices & 800 & 5000 & 20000 & 125000 & 500000 \\
\hline
edges & 1160 & 7400 & 29800 & 187000 & 749000 \\
\hline
time [s] & 0.02 & 0.03 & 0.2 & 0.6 & 2.5 \\
\hline
\end{tabular} 
\caption{\label{tab:saucy}Number of vertices and edges of the colored graphs, the runtime of \textsc{Saucy}, and the number of (super-)features of the social network model without and with 10\% evidence.}
\end{table}
\begin{figure}[t!]
\begin{center}
\includegraphics[width=0.482\textwidth]{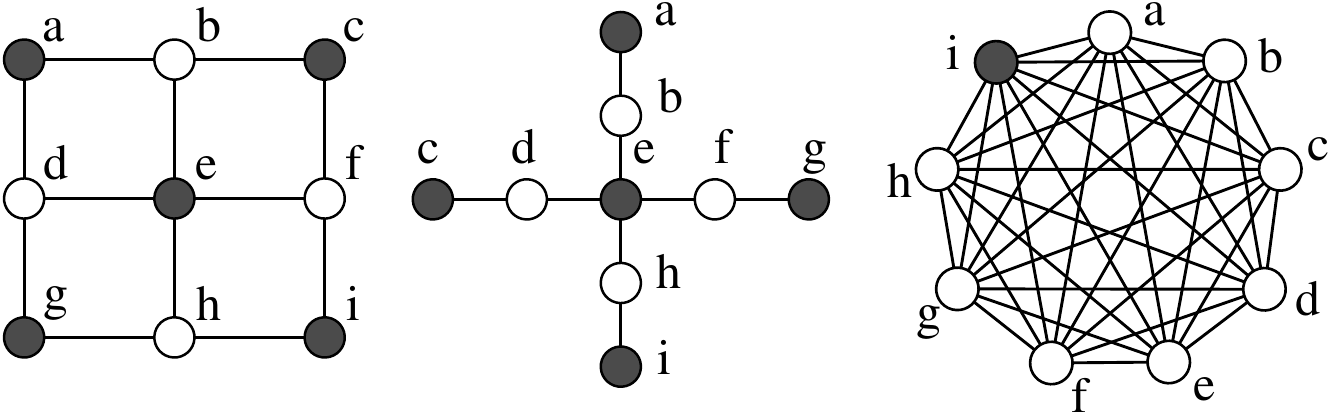}
\caption{\label{fig:models}From left to right: the \emph{$3$-grid}, the \emph{$3$-connected cliques}, and the \emph{$3$-complete graph} models. }
\end{center}
\end{figure}

\section{Experiments}

Two graphical models were used to evaluate the symmetry detection approach. The ``Friends \& Smokers" Markov logic network where for a random 10\% of all people it is known (a) whether they smoke or not and (b) who 10 of their friends are~\cite{singla2008lifted}. Moreover, we used the $k\mathsf{x}k$ grid model, an established and well-motivated lattice model with numerous applications~\cite{restrepo:2011}. All experiments were conducted on a PC with an AMD Athlon dual core 5400B 1.0 GHz processor and 3 GB RAM.  Table~\ref{tab:saucy} lists the results for varying model sizes. \textsc{Saucy}'s runtime scales roughly quadratic with the number of vertices and it performs better for the $k\mathsf{x}k$ grid models. This might be due to the larger sets of generators for the permutation groups of the social network model. Table~\ref{tab:saucy} also lists the number of features of the ground social network model (features), the number of feature orbits without (orbs w/o) and with (orbs w/) 10\% evidence.

We proceeded to compare the performance of the orbital Markov chains with state-of-the-art  algorithms for sampling independent sets. We used \textsc{Gap}\cite{GAP4}, a system for computational discrete algebra, and the \textsc{Orb} package\cite{mueller:2007}\footnote{http://www.gap-system.org/Packages/orb.html} to implement the sampling algorithms. The experiments can easily be replicated by installing \textsc{Gap} and the \textsc{Orb} package and by running the  \textsc{Gap} files available at a dedicated code repository\footnote{http://code.google.com/p/lifted-mcmc/}. For the evaluation of the sampling algorithms we selected three different graph topologies exhibiting varying degrees of symmetry:

\begin{figure}[t!]
\centering
\includegraphics[width=0.5\textwidth]{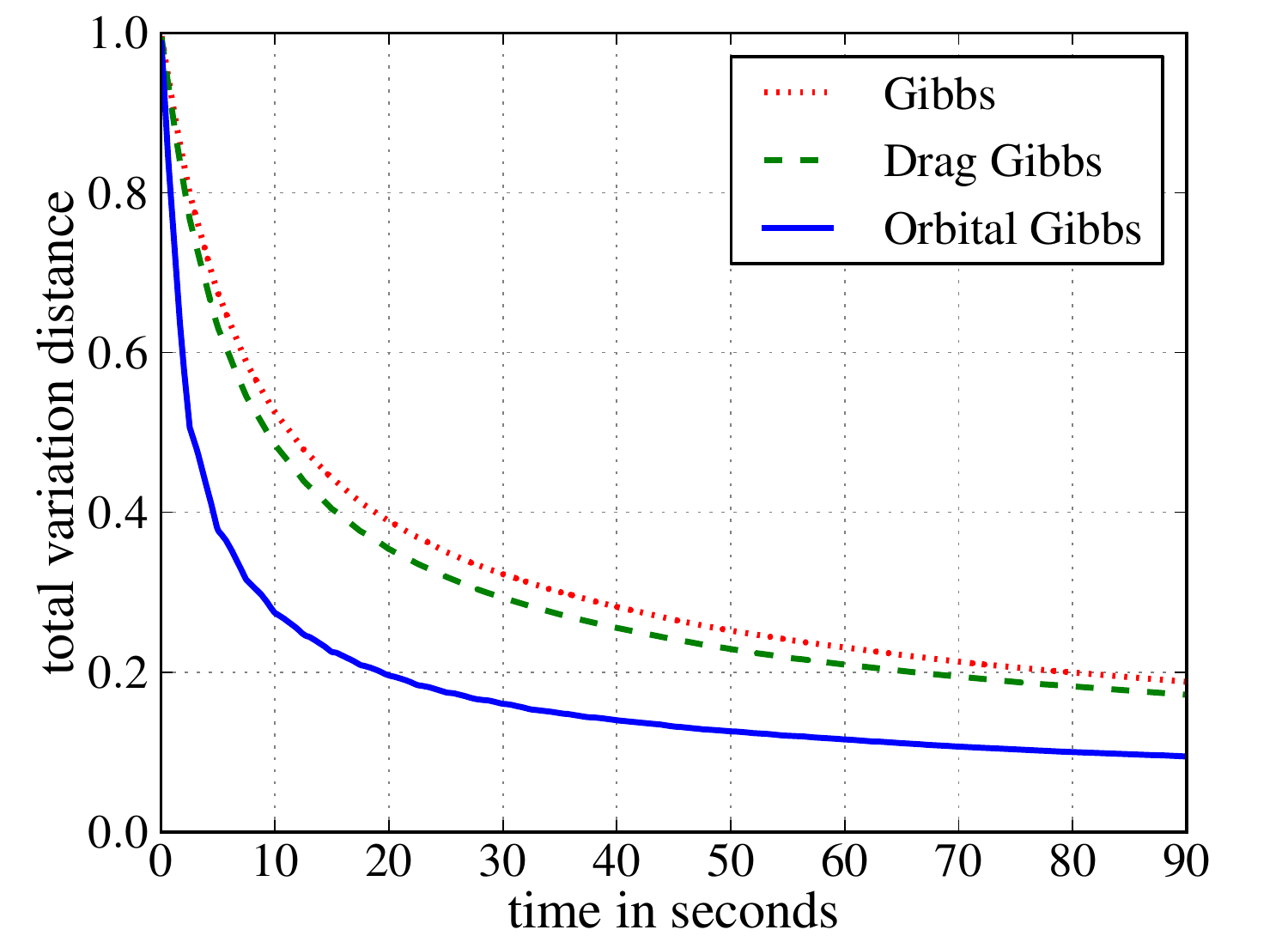}
\includegraphics[width=0.5\textwidth]{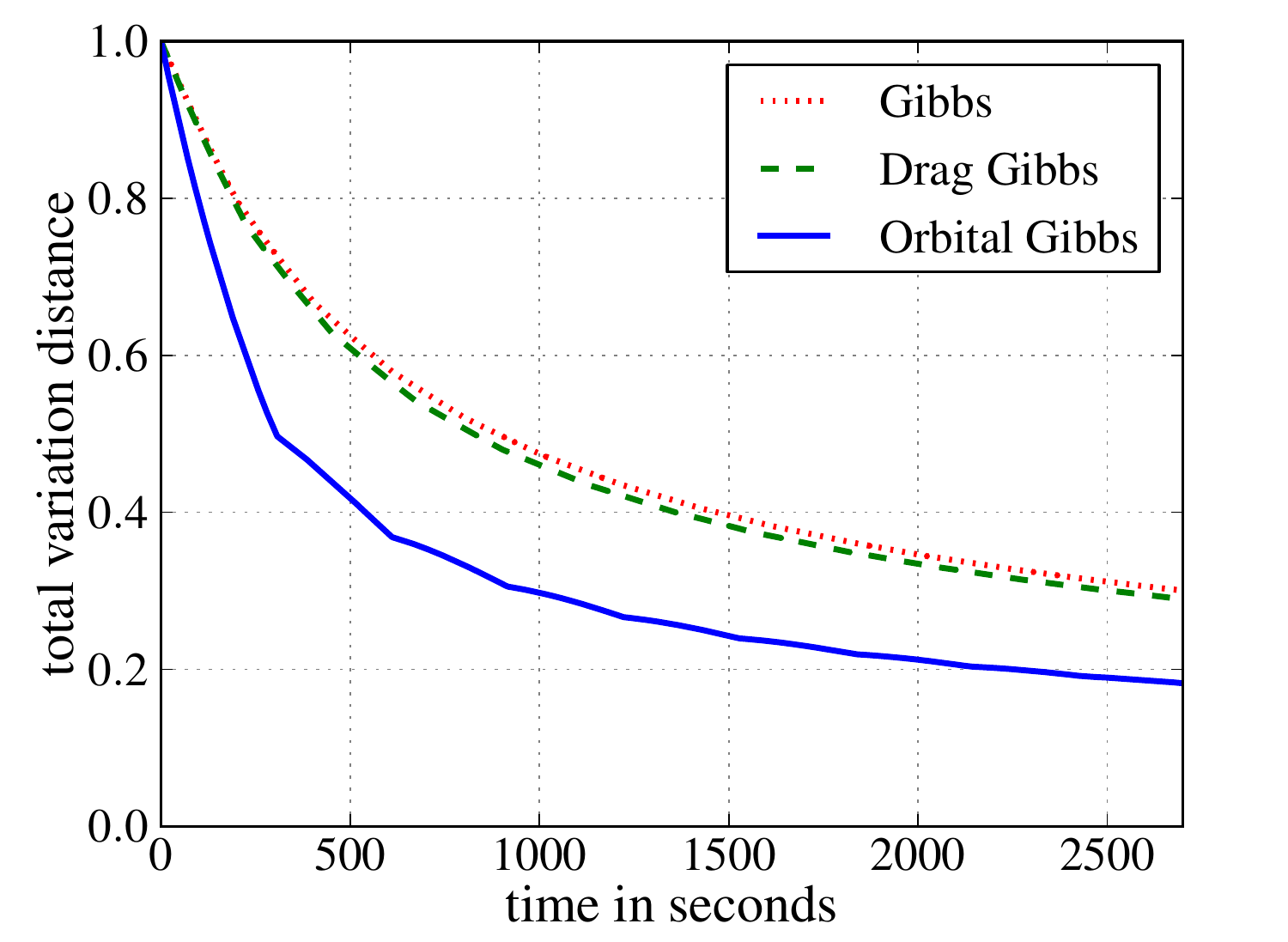}
\caption{\label{fig:hm-plot}The results of the three Gibbs samplers for the \emph{$5$-grid model} (top) and the \emph{$6$-grid model} (bottom).} 
\end{figure}

The \emph{$k$-grid model} is the $2$-dimensional $k\mathsf{x}k$ grid. An instance of the model for $k=3$ is depicted  in Figure~\ref{fig:models} (left). Here, the generating set of the permutation group $\mathfrak{G}$ computed by \textsc{Saucy} is $\{(a\ c)(d\ f)(g\ i), (a\ i)(b\ f)(d\ h)\}$ and $|\mathfrak{G}| = 8$. The permutation group  $\mathfrak{G}$ partitions the set $\{0, 1\}^9$ in $102$ orbits with each orbit having a cardinality in $\{1, 2, 4, 8\}$. 

The \emph{$k$-connected cliques model} is a graph with $k+1$ distinct cliques each of size $k-1$ and each connected with one edge to the same vertex. Statistical relational formalisms such as Markov logic networks often lead to similar graph topologies. An instance for $k=3$ is depicted in Figure~\ref{fig:models} (center).  Here, the generating set of $\mathfrak{G}$ computed by \textsc{Saucy} is $\{(a\ g)(b\ f), (a\ c)(b\ d), (a\ i)(b\ h)\}$ and $|\mathfrak{G}| = 24$. The permutation group  $\mathfrak{G}$ partitions the set $\{0, 1\}^9$ in $70$ orbits with cardinalities in $\{1, 4, 6, 12, 24\}$.  

The \emph{$k$-complete graph model} is a complete graph with $k^2$ vertices.  Figure~\ref{fig:models} (right) depicts an instance for $k=3$. Here, the generating set of $\mathfrak{G}$ computed by \textsc{Saucy} is $\{(b\ c), (b\ d), (b\ e), (b\ f), (b\ g), (b\ h), (b\ i), (a\ b)\}$ and $|\mathfrak{G}| = 9! = 362880$. The permutation group  $\mathfrak{G}$ partitions the set $\{0, 1\}^9$ in $10$ orbits with each orbit having a cardinality in $\{1, 9, 36, 84, 126\}$.  

\begin{figure}[t!]
\begin{center}
\includegraphics[width=0.5\textwidth]{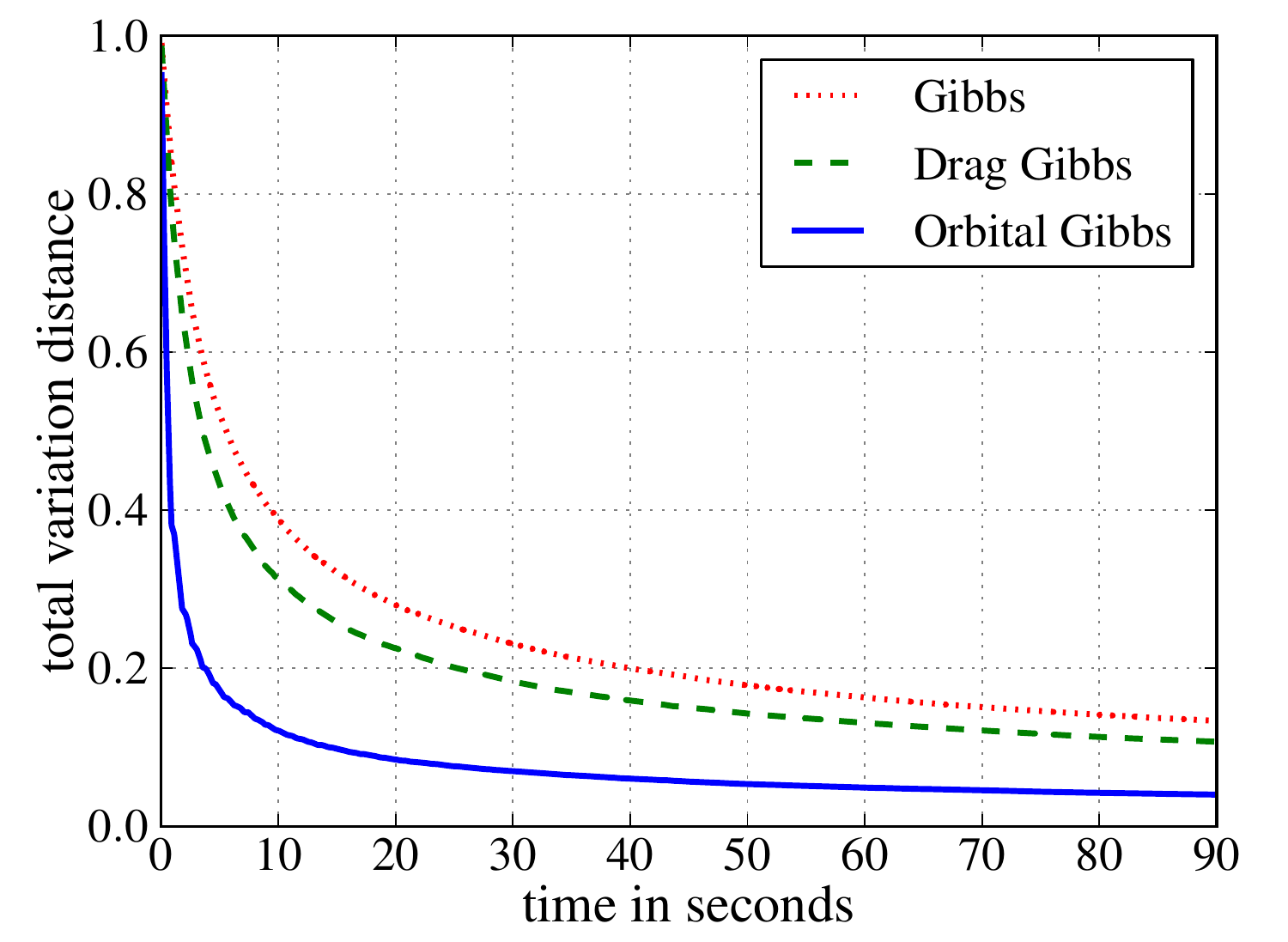}
\caption{\label{fig:mlnmodel-plot}The results of the three Gibbs samplers for the  \emph{$5$-connected cliques} model.}
\end{center}
\end{figure}

\begin{figure}[t!]
\begin{center}
\includegraphics[width=0.5\textwidth]{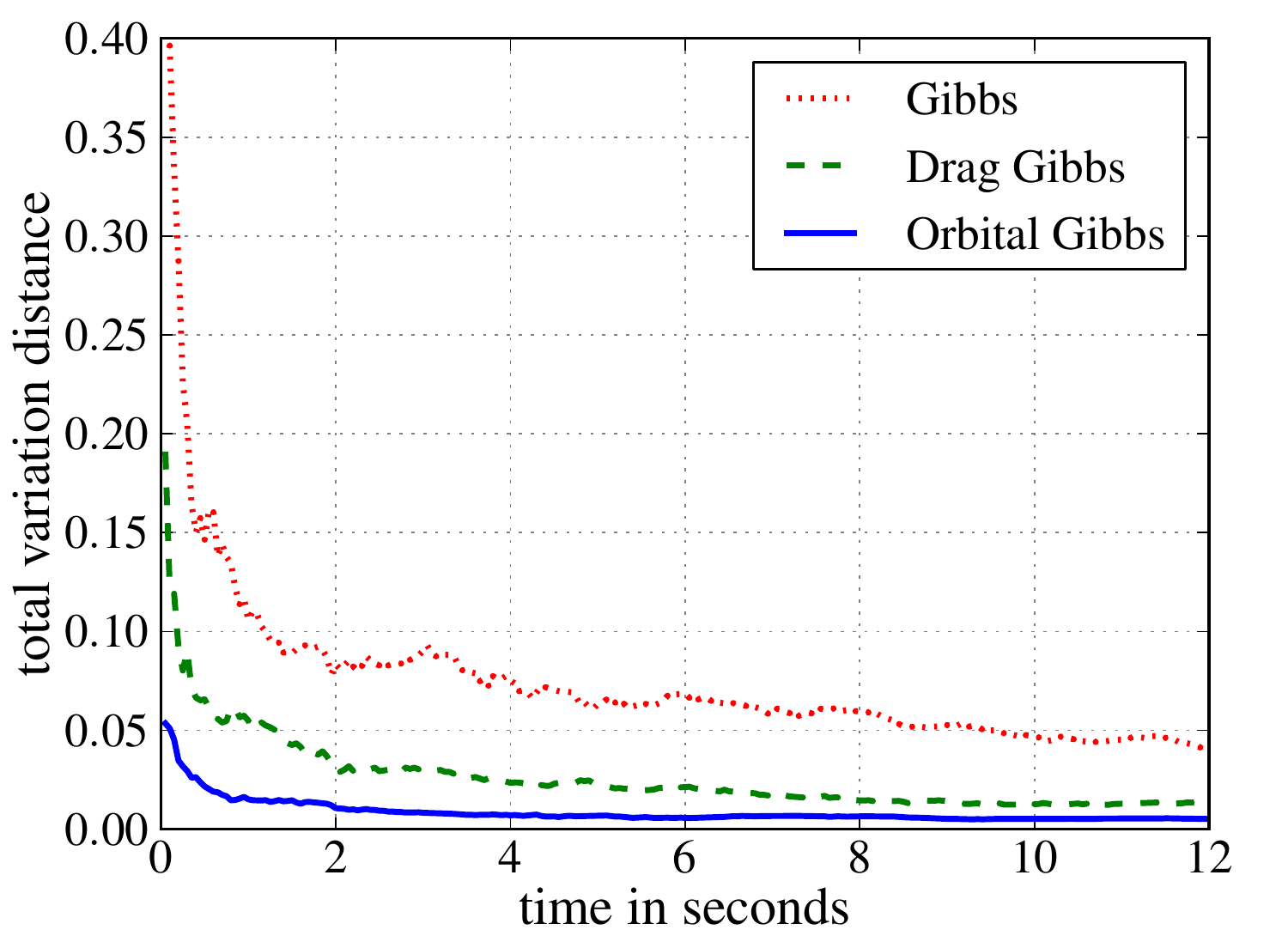}
\caption{\label{fig:symmodel-plot}The results of the three Gibbs samplers for the  \emph{$5$-complete graph} model.}
\end{center}
\end{figure}

\textsc{Saucy} needed at most 5 ms to compute the sets of generators for the permutation groups of the three models for $k = 6$. We generated samples of the probability measure $\pi_{\lambda}$ on $\mathcal{I}(G)$ for $\lambda = 1$  and the three different graph topologies by running (a) the \emph{insert/delete} chain, (b) the \emph{insert/delete/drag} chain \cite{dyer2000markov}, and (c) the orbital \emph{insert/delete} chain. Each chain was started in the state corresponding to the empty set and no burn-in period was used.
The orbital \emph{insert/delete} chain did not require more RAM and needed 50 microseconds per sample which amounts to an overhead of about 25\% relative to the 40 microseconds of the \emph{insert/delete} chain. The 25\% overhead remained constant and independent of the size of the graphs. Since the sampling algorithms create large files with all accumulated samples, I/O overhead is included in these times.
For each of the three topologies and each of the three Gibbs samplers, we computed the total variation distance between the distribution approximated using \emph{all} accumulated samples and the true distribution $\pi_1$. Figure~\ref{fig:hm-plot} plots the total variation distance over elapsed time for the \emph{$k$-grid model} for $k=5$ and $k=6$. The orbital \emph{insert/delete} chain (Orbital Gibbs) converges the fastest. The \emph{insert/delete/drag} chain (Drag Gibbs) converges faster than the \emph{insert/delete} chain (Gibbs) but since there is a small computational overhead of the \emph{insert/delete/drag} chain the difference is less pronounced for $k=6$. The same results are observable for the other graph topologies (see Figures~\ref{fig:mlnmodel-plot} and \ref{fig:symmodel-plot}) where the orbital Markov chain outperforms the others. In summary, the larger the cardinalities of the orbits induced by the symmetries the faster converges the orbital Gibbs sampler  relative to the other chains.

\section{Discussion} 
 
 The mindful reader might have recognized a similarity to \emph{lumping} of Markov chains which amounts to partitioning the state space of the chain~\cite{buchholz:1994}. Computing the coarsest lumping quotient of a
 Markov chain with a bi-simulation procedure is linear in the \emph{number of non-zero probability transitions} of the chain and, hence, in most cases exponential in the number of random variables. Since merely counting equivalence classes in the P\'{o}lya theory setting is a $\#$P-complete problem~\cite{goldberg:2001} there are clear computational limitations to this approach.
Orbital Markov chains, on the other hand, combine the advantages of a compact representation of symmetries as generating sets of  permutation groups with highly efficient product replacement algorithms and, therefore, provide the advantages of \emph{lumping} while avoiding the intractable explicit computation of the partition of the state space.
 
One can apply orbital Markov chains to other graphical models that exhibit symmetries such as the Ising model.
Since Markov chains in general and Gibbs samplers in particular are components in numerous algorithms (cf.~\cite{stephens:2000,olle:2002,RD:2006,gomes2007sampling,liang:2007,blei:2003}), we expect orbital Markov chains to improve the algorithms' performance  when applied to problems that exhibit symmetries. For instance, sampling algorithms for statistical relational languages are obvious candidates for improvement. Future work will include the integration of orbital Markov chains with algorithms for  marginal as well as maximum a-posteriori inference. We will also apply the symmetry detection approach to make existing inference algorithms more efficient by, for instance, using symmetry breaking constraints in combinatorial optimization approaches to maximum a-posteriori inference in Markov logic networks (cf. \cite{riedel:2008,niepert2010uai,niepert2011log}).

While we have shown that permutation groups are computable with graph automorphism algorithms for a large class of models it is also possible to \emph{assume} certain symmetries in the model in the same way (conditional) independencies are assumed in the design stage of a probabilistic graphical model. Orbital Markov chains could easily incorporate these symmetries in form of permutation groups.

\subsubsection*{Acknowledgments} 
 
Many thanks to Guy Van den Broeck, Kristian Kersting, Martin Mladenov, and Babak Ahmadi for insightful discussions concerning lifted inference, to J\"{u}rgen M\"{u}ller for helpful remarks on the product replacement algorithm, and to all those who have contributed to the \textsc{Gap} system and the \textsc{Orb} package.
 
\bibliographystyle{abbrv}
\bibliography{uai-2012}

\begin{thebibliography}{10}

\bibitem{aldous:1983}
D.~Aldous.
\newblock Random walks on finite groups and rapidly mixing {M}arkov chains.
\newblock In {\em Seminaire de Probabilites XVII}, pages 243--297. 1983.

\bibitem{aloul:2003}
F.~A. Aloul, K.~A. Sakallah, and I.~L. Markov.
\newblock Efficient symmetry breaking for boolean satisfiability.
\newblock In {\em Proceedings of the 18th International Joint Conference on
  Artificial Intelligence}, pages 271--276, 2003.

\bibitem{blei:2003}
D.~M. Blei, A.~Y. Ng, and M.~I. Jordan.
\newblock Latent dirichlet allocation.
\newblock {\em J. Mach. Learn. Res.}, 3:993--1022, Mar. 2003.

\bibitem{bubley:1997}
R.~Bubley and M.~Dyer.
\newblock Path coupling: A technique for proving rapid mixing in markov chains.
\newblock In {\em Proceedings of the 38th Symposium on Foundations of Computer
  Science}, pages 223--231, 1997.

\bibitem{buchholz:1994}
P.~Buchholz.
\newblock Exact and ordinary lumpability in finite markov chains.
\newblock {\em Journal of Applied Probability}, 31(1):pp. 59--75, 1994.

\bibitem{celler:1995}
F.~Celler, C.~R. Leedham-Green, S.~H. Murray, A.~C. Niemeyer, and E.~O'brien.
\newblock Generating random elements of a finite group.
\newblock {\em Communications in Algebra}, 23(13):4931--4948, 1995.

\bibitem{crawford:1992}
J.~Crawford.
\newblock A theoretical analysis of reasoning by symmetry in first-order logic.
\newblock In {\em Proceedings of the Workshop on Tractable Reasoning}, 1992.

\bibitem{darga:2008}
P.~T. Darga, K.~A. Sakallah, and I.~L. Markov.
\newblock Faster symmetry discovery using sparsity of symmetries.
\newblock In {\em Proceedings of the 45th annual Design Automation Conference},
  pages 149--154, 2008.

\bibitem{braz:2005}
R.~de~Salvo~Braz, E.~Amir, and D.~Roth.
\newblock Lifted first-order probabilistic inference.
\newblock In {\em Proceedings of the 19th International Joint Conference on
  Artificial Intelligence}, pages 1319--1325, 2005.

\bibitem{diaconis:1980}
P.~Diaconis and D.~Freedman.
\newblock De finetti's generalizations of exchangeability.
\newblock In {\em Studies in Inductive Logic and Probability}, volume~II. 1980.

\bibitem{drescher:2011}
C.~Drescher, O.~Tifrea, and T.~Walsh.
\newblock Symmetry-breaking answer set solving.
\newblock {\em AI Commun.}, 24(2):177--194, 2011.

\bibitem{dyer:1998}
M.~Dyer and C.~Greenhill.
\newblock A more rapidly mixing markov chain for graph colorings.
\newblock {\em Random Struct. Algorithms}, 13(3-4):285--317, 1998.

\bibitem{dyer2000markov}
M.~Dyer and C.~Greenhill.
\newblock On markov chains for independent sets.
\newblock {\em Journal of Algorithms}, 35(1):17--49, 2000.

\bibitem{finetti:1972}
B.~Finetti.
\newblock {\em Probability, induction and statistics: the art of guessing}.
\newblock Probability and mathematical statistics. Wiley, 1972.

\bibitem{GAP4}
The GAP~Group.
\newblock {\em {GAP -- Groups, Algorithms, and Programming, Version 4.4.12}},
  2008.

\bibitem{gogate2011probabilistic}
V.~Gogate and P.~Domingos.
\newblock Probabilistic theorem proving.
\newblock In {\em Proceedings of the 27th Conference on Uncertainty in
  Artificial Intelligence}, pages 256--265, 2011.

\bibitem{gogate:2012}
V.~Gogate, A.~Jha, and D.~Venugopal.
\newblock Advances in lifted importance sampling.
\newblock In {\em Proceedings of the 26th Conference on Artificial
  Intelligence}, 2012.

\bibitem{goldberg:2001}
L.~A. Goldberg.
\newblock Computation in permutation groups: counting and randomly sampling
  orbits.
\newblock In {\em Surveys in Combinatorics}, pages 109--143. Cambridge
  University Press, 2001.

\bibitem{gomes2007sampling}
C.~Gomes, J.~Hoffmann, A.~Sabharwal, and B.~Selman.
\newblock From sampling to model counting.
\newblock In {\em Proceedings of the 20th International Joint Conference on
  Artificial Intelligence}, pages 2293--2299, 2007.

\bibitem{olle:2002}
O.~H\"{a}ggstr\"{o}m.
\newblock {\em {Finite Markov Chains and Algorithmic Applications}}.
\newblock London Mathematical Society Student Texts. Cambridge University
  Press, 2002.

\bibitem{jaeger:2012}
M.~Jaeger.
\newblock Lower complexity bounds for lifted inference.
\newblock {\em CoRR}, abs/1204.3255, 2012.

\bibitem{kersting2009counting}
K.~Kersting, B.~Ahmadi, and S.~Natarajan.
\newblock Counting belief propagation.
\newblock In {\em Proceedings of the 25th Conference on Uncertainty in
  Artificial Intelligence}, pages 277--284, 2009.

\bibitem{liang:2007}
P.~Liang, M.~I. Jordan, and B.~Taskar.
\newblock A permutation-augmented sampler for dp mixture models.
\newblock In {\em Proceedings of the 24th International Conference on Machine
  Learning}, pages 545--552, 2007.

\bibitem{luby:1999}
M.~Luby and E.~Vigoda.
\newblock Fast convergence of the glauber dynamics for sampling independent
  sets.
\newblock {\em Random Struct. Algorithms}, 15(3-4):229--241, 1999.

\bibitem{luks:1982}
E.~M. Luks.
\newblock Isomorphism of graphs of bounded valence can be tested in polynomial
  time.
\newblock {\em J. Computer and System Sciences}, 25(1):42--65, 1982.

\bibitem{Margot:2003}
F.~Margot.
\newblock Exploiting orbits in symmetric ilp.
\newblock {\em Math. Program.}, 98(1-3):3--21, 2003.

\bibitem{Margot:2010}
F.~Margot.
\newblock Symmetry in integer linear programming.
\newblock In {\em 50 Years of Integer Programming 1958-2008}, pages 647--686.
  Springer-Verlag, 2010.

\bibitem{mckay:1981}
B.~D. McKay.
\newblock Practical graph isomorphism.
\newblock {\em Congressus Numerantium}, 30:45--87, 1981.

\bibitem{milch:2006}
B.~Milch and S.~J. Russell.
\newblock General-purpose mcmc inference over relational structures.
\newblock In {\em Proceedings of the 22nd Conference in Uncertainty in
  Artificial Intelligence}, pages 349--358, 2006.

\bibitem{mladenov12aistats}
M.~Mladenov, B.~Ahmadi, and K.~Kersting.
\newblock Lifted linear programming.
\newblock In {\em Proceedings of the 15th International Conference on
  Artificial Intelligence and Statistics}, 2012.

\bibitem{mueller:2007}
J.~M\"{u}ller, M.~Neunh\"{o}ffer, and R.~Wilson.
\newblock Enumerating big orbits and an application: $b$ acting on the cosets
  of $fi_{23}$.
\newblock {\em Journal of Algebra}, 314(1):75--96, 2007.

\bibitem{niepert2010uai}
M.~Niepert.
\newblock A delayed column generation strategy for exact $k$-bounded map
  inference in markov logic networks.
\newblock In {\em Proceedings of the 25th Conference on Uncertainty in
  Artificial Intelligence}, pages 384--391, 2010.

\bibitem{niepert2011log}
M.~Niepert, J.~Noessner, and H.~Stuckenschmidt.
\newblock Log-linear description logics.
\newblock In {\em Proceedings of the 22nd International Joint Conference on
  Artificial Intelligence}, pages 2153--2158, 2011.

\bibitem{ostrowski:2011}
J.~Ostrowski, J.~Linderoth, F.~Rossi, and S.~Smriglio.
\newblock Orbital branching.
\newblock {\em Math. Program.}, 126(1):147--178, 2011.

\bibitem{pak:2000}
I.~Pak.
\newblock The product replacement algorithm is polynomial.
\newblock In {\em Proceedings of the 41st Annual Symposium on Foundations of
  Computer Science}, pages 476--485, 2000.

\bibitem{poole:2003}
D.~Poole.
\newblock First-order probabilistic inference.
\newblock In {\em Proceedings of the 18th Joint Conference on Artificial
  Intelligence}, pages 985--991, 2003.

\bibitem{restrepo:2011}
R.~Restrepo, J.~Shin, P.~Tetali, E.~Vigoda, and L.~Yang.
\newblock Improved mixing condition on the grid for counting and sampling
  independent sets.
\newblock In {\em Proceedings of the 52nd Symposium on Foundations of Computer
  Science}, pages 140--149, 2011.

\bibitem{RD:2006}
M.~Richardson and P.~Domingos.
\newblock Markov logic networks.
\newblock {\em Machine Learning}, 62(1-2), 2006.

\bibitem{riedel:2008}
S.~Riedel.
\newblock Improving the accuracy and efficiency of map inference for markov
  logic.
\newblock In {\em Proceedings of the Conference on Uncertainty in Artificial
  Intelligence}, pages 468--475, 2008.

\bibitem{sen:2009}
P.~Sen, A.~Deshpande, and L.~Getoor.
\newblock Bisimulation-based approximate lifted inference.
\newblock In {\em Proceedings of the 25th Conference on Uncertainty in
  Artificial Intelligence}, 2009.

\bibitem{singla2008lifted}
P.~Singla and P.~Domingos.
\newblock Lifted first-order belief propagation.
\newblock In {\em Proceedings of the 23rd Conference on Artificial
  Intelligence}, pages 1094--1099, 2008.

\bibitem{sly:2010}
A.~Sly.
\newblock Computational transition at the uniqueness threshold.
\newblock In {\em Proceedings of the 51st Symposium on Foundations of Computer
  Science}, pages 287--296, 2010.

\bibitem{stephens:2000}
M.~Stephens.
\newblock Dealing with label switching in mixture models.
\newblock {\em Journal of the Royal Statistical Society Series B},
  62(4):795--809, 2000.

\bibitem{broeck:2011}
G.~Van~den Broeck.
\newblock On the completeness of first-order knowledge compilation for lifted
  probabilistic inference.
\newblock In {\em Advances in Neural Information Processing Systems}, pages
  1386--1394, 2011.

\bibitem{weitz:2006}
D.~Weitz.
\newblock Counting independent sets up to the tree threshold.
\newblock In {\em Proceedings of the 38th ACM Symposium on Theory of
  computing}, pages 140--149, 2006.

\end{thebibliography}

\appendix

\newpage 
\section{Proof of Theorem~\ref{theorem-orbital}}

We first prove (a). Since $\mathcal{M'}$ is aperiodic we have, for each state $x \in \Omega$ and every time step $t \geq 0$, a non-zero probability for the Markov chain $\mathcal{M}'$ to remain in state $x$ at time $t+1$. At each time $t+1$, the orbital Markov chain transitions uniformly at random to one of the states in the orbit of the original chain's state at time $t+1$. Since every state is an element of its own orbit, we have, for every state $x \in \Omega$ and every time step $t \geq 0$, a non-zero probability for the Markov chain $\mathcal{M}$ to remain in state $x$ at time $t+1$. Hence, $\mathcal{M}$ is aperiodic. The proof of statement (b) is accomplished in an analogous fashion and omitted.

Let $P(x, y)$ and $P'(x, y)$ be the probabilities of $\mathcal{M}$ and $\mathcal{M'}$, respectively, to transition from state $x$ to state $y$. Since $\pi$ is a reversible distribution for $\mathcal{M'}$ we have that $\pi(x)P'(x, y) = \pi(y)P'(y, x)$ for all states $x, y \in \Omega$. For every state $x \in \Omega$ let $x^{\mathfrak{G}}$ be the orbit of $x$.
Let $\mathfrak{G}_x := \{ \mathfrak{g}\in \mathfrak{G}\ |\ x^{\mathfrak{g}} = x\}$ 
be the stabilizer subgroup of $x$ with respect to $\mathfrak{G}$. We have that 
\vspace{-1mm}
\begin{equation}
\vspace{-1mm}
\begin{split}
 \sum_{\mathfrak{g} \in \mathfrak{G}}P'(x, y^{\mathfrak{g}})  & = \sum_{y' \in y^{\mathfrak{G}}}|\mathfrak{G}_{y'}|P'(x, y')  \\
& = |\mathfrak{G}_{y}|\sum_{y' \in y^{\mathfrak{G}}}P'(x, y') \\
& = (|\mathfrak{G}|/|y^{\mathfrak{G}}|)\sum_{y' \in y^{\mathfrak{G}}}P'(x, y') 
\end{split}
\end{equation}
where the last two equalities follow from the orbit-stabilizer theorem.
We will now prove that $\pi(x)P(x, y) = \pi(y)P(y, x)$ for all states $x, y \in \Omega$. By definition of the orbital Markov chain we have that $\pi(x)P(x, y) = \pi(x)(1/|y^{\mathfrak{G}}|) \sum_{y'\in y^{\mathfrak{G}}}P'(x, y')$ and, by equation (1), $\pi(x)(1/|y^{\mathfrak{G}}|) \sum_{y'\in y^{\mathfrak{G}}}P'(x, y')$
\vspace{-1mm}
\begin{equation*}
\vspace{-2mm}
\begin{split}
 & =  \pi(x)(1/|y^{\mathfrak{G}}|)(|y^{\mathfrak{G}}|/|\mathfrak{G}|)\sum_{\mathfrak{g} \in \mathfrak{G}}P'(x, y^{\mathfrak{g}})  \\
 & = \pi(x) (1/|\mathfrak{G}|) \sum_{\mathfrak{g} \in \mathfrak{G}}P'(x, y^{\mathfrak{g}})   \\
 & =  (1/|\mathfrak{G}|) \sum_{\mathfrak{g} \in \mathfrak{G}} \pi(x)P'(x, y^{\mathfrak{g}}).  
\end{split}
\end{equation*}
Since $P'$ is reversible and $\pi(x)= \pi(x^{\mathfrak{g}})$ for all $\mathfrak{g} \in \mathfrak{G}$ we have $(1/|\mathfrak{G}|) \sum_{\mathfrak{g} \in \mathfrak{G}} \pi(x)P'(x, y^{\mathfrak{g}}) = (1/|\mathfrak{G}|) \sum_{\mathfrak{g} \in \mathfrak{G}} \pi(y^{\mathfrak{g}})P'(y^{\mathfrak{g}}, x) = \pi(y) (1/|\mathfrak{G}|) \sum_{\mathfrak{g} \in \mathfrak{G}} P'(y^{\mathfrak{g}}, x)$.
Now, since $P'(x, y)=P'(x^{\mathfrak{g}}, y^{\mathfrak{g}})$ for all $x, y\in\Omega$ and all $\mathfrak{g}\in \mathfrak{G}$ by assumption, we have that 
$\pi(y) (1/|\mathfrak{G}|) \sum_{\mathfrak{g} \in \mathfrak{G}} P'(y^{\mathfrak{g}}, x) = \pi(y) (1/|\mathfrak{G}|) \sum_{\mathfrak{g} \in \mathfrak{G}} P'(y, x^{-\mathfrak{g}}) = \pi(y) (1/|\mathfrak{G}|) \sum_{\mathfrak{g} \in \mathfrak{G}} P'(y, x^{\mathfrak{g}})$ and, again by equation (1), $\pi(y) (1/|\mathfrak{G}|) \sum_{\mathfrak{g} \in \mathfrak{G}} P'(y, x^{\mathfrak{g}})$
\vspace{-1mm}
\begin{equation*}
\vspace{-1mm}
\begin{split}
\hspace{5mm}
& = \pi(y) (1/|\mathfrak{G}|) (|\mathfrak{G}|/|x^{\mathfrak{G}}|)\sum_{x' \in x^{\mathfrak{G}}}P'(y, x') \\
& =  \pi(y) (1/|x^{\mathfrak{G}}|)\sum_{x' \in x^{\mathfrak{G}}}P'(y, x') =\pi(y)P(y, x). 
\hspace{5mm}\hfill\square
\end{split}
\end{equation*}

\section{Proof of Theorem~\ref{theorem-rho}}

Let $H: \Omega \times \Omega \rightarrow \mathbb{N}$ be the Hamming distance between any two elements in $\Omega$. We provide a path coupling argument on the set of pairs having Hamming distance $1$.
Let $X$ and $Y$ be two independent sets which differ only at one vertex $v$ with degree $d$. We assume, without loss of generality, that $v \in X \setminus Y$.
Choose a vertex $w$ uniformly at random. We distinguish five cases:
\begin{enumerate}
\item[(i)] if $w=v$ then
sample one $\mathfrak{g} \in \mathfrak{G}$ uniformly at random and let $(X',Y')=(X^{\mathfrak{g}},X^{\mathfrak{g}})$ with probability $\lambda/(1 + \lambda)$, otherwise let $(X',Y')=(Y^{\mathfrak{g}},Y^{\mathfrak{g}})$; Hence, $H(X',Y') = 0$ with probability $1$.
\item[(ii)] if $w \neq v$ and $w \in X$ then
sample one $\mathfrak{g} \in \mathfrak{G}$ uniformly at random and let $(X',Y')=((X\setminus\{w\})^{\mathfrak{g}},(Y\setminus\{w\})^{\mathfrak{g}})$ with probability $1/(1 + \lambda)$, otherwise let $(X',Y')=(X^{\mathfrak{g}},Y^{\mathfrak{g}})$. In both cases, we have that $H(X',Y') = 1$.
\item[(iii)] if $w \neq v$, $w \notin X$ and $w$ has no neighbor in $X$ then
sample one $\mathfrak{g} \in \mathfrak{G}$ uniformly at random and let $(X',Y')=((X\cup\{w\})^{\mathfrak{g}},(Y\cup\{w\})^{\mathfrak{g}})$ with probability $\lambda/(1 + \lambda)$, otherwise let $(X',Y')=(X^{\mathfrak{g}},Y^{\mathfrak{g}})$; In both cases, we have that $H(X',Y') = 1$.
\item[(iv)] if $w \neq v$, $w \notin X$ and $w$ has a neighbor in $X$ but not in $Y,$ then sample one $\mathfrak{g} \in \mathfrak{G}$ uniformly at random. Let $\mathfrak{G'} := \{\mathfrak{g}' \in \mathfrak{G}\ |\ X^{\mathfrak{g}} = (Y\cup\{w\})^{\mathfrak{g}'}\}$. If $\mathfrak{G'} \neq \emptyset$ then sample one  $g' \in \mathfrak{G'}$ uniformly at random and let $(X',Y')=(X^{\mathfrak{g}}, (Y\cup\{w\})^{\mathfrak{g}'})$ with probability $\lambda/(1 + \lambda)$. In this case we have $H(X',Y')=0$. If $\mathfrak{G'} = \emptyset$ then let $(X',Y')=({X}^{\mathfrak{g}}, (Y\cup\{w\})^{\mathfrak{g}})$ with probability $\lambda/(1 + \lambda)$. Here we have $H(X',Y') = 2$. Otherwise let $(X',Y')=(X^{\mathfrak{g}}, Y^{\mathfrak{g}})$. Here, we have $H(X',Y')=1$.
\item[(v)] in all other cases sample one $\mathfrak{g} \in \mathfrak{G}$ uniformly at random and let $(X',Y')=(X^{\mathfrak{g}}, Y^{\mathfrak{g}})$. Here we have with probability $1$ that $H(X',Y')=1$.
\end{enumerate}
In summary, we have that $$\mathbf{E}[H(X',Y')-1] \leq -\frac{1}{n} + \varrho (2\rho - 1)\frac{\lambda}{(1+\lambda)}$$  where $\varrho = \Pr[w \neq v$, $w \notin X$, and $w$ has a neighbor in $X$ but not in $Y]$ and $\rho = \Pr[X \notin (Y\cup\{w\})^\mathfrak{G}\ |\ w \neq v$, $w \notin X$, and $w$ has a neighbor in $X$ but not in $Y]$. If $\rho \leq 0.5$ we have that $\mathbf{E}[H(X',Y')-1] \leq -\frac{1}{n}$ otherwise we have that $\ \ \mathbf{E}[H(X',Y')-1] \leq$ $$ -\frac{1}{n} + \frac{d}{n}(2\rho -1) \frac{\lambda}{(1+\lambda)} \leq -\frac{1}{n} + \frac{\Delta}{n}(2\rho -1) \frac{\lambda}{(1+\lambda)} .$$ 
Hence, $\mathcal{M}(\mathcal{I}(G))$ mixes rapidly if either $\rho \leq 0.5$ or $\lambda((2\rho -1) \Delta  - 1) < 1$. 
For $\lambda((2\rho -1)\Delta  - 1) = 1$ 	one can verify that there exists an $\alpha > 0$ such that $\Pr[H(X_{t+1},Y_{t+1}) \neq H(X_t, Y_t)] \geq \alpha$ for all $t$. $\hfill\square$

\end{document}